\documentclass[conference]{IEEEtran}
\IEEEoverridecommandlockouts

\usepackage{cite}
\usepackage{amsmath,amssymb,amsfonts}
\usepackage{booktabs}
\usepackage{multirow}
\usepackage{array}
\usepackage{tabularx}
\usepackage{algorithm}
\usepackage{algpseudocode}
\usepackage{graphicx}
\usepackage{textcomp}
\usepackage{xcolor}
\usepackage{comment}
\usepackage{bm}
\usepackage{color}
\usepackage{float}
\usepackage{hyperref}

\def\BibTeX{{\rm B\kern-.05em{\sc i\kern-.025em b}\kern-.08em
    T\kern-.1667em\lower.7ex\hbox{E}\kern-.125emX}}

\begin{document}
\title{3D CT-to-PET Translation via \\ Latent
Brownian Bridge Diffusion\\

\thanks{}
}

\author{\IEEEauthorblockN{Sarita Mourya\IEEEauthorrefmark{1}\textsuperscript{1}, Francesco Di Feola\IEEEauthorrefmark{2}\textsuperscript{1}, Pierangelo Veltri\IEEEauthorrefmark{3}, Paolo Soda\IEEEauthorrefmark{2}\IEEEauthorrefmark{4}}
\IEEEauthorblockA{\IEEEauthorrefmark{2}Department of Diagnostics and Intervention, Radiation Physics, Biomedical Engineering, Umeå University, Sweden}
\IEEEauthorblockA{\IEEEauthorrefmark{4} Unit of Artificial Intelligence and Computer Systems, Department of Engineering, Università Campus Bio-Medico di Roma}
\IEEEauthorblockA{\IEEEauthorrefmark{1} Department of Medical and Surgical Sciences, Università "Magna Græcia" di Catanzaro
Catanzaro, Italy}
\IEEEauthorblockA{\IEEEauthorrefmark{3} Department of Computer Science, Modeling, Electronics and Systems Engineering (DIMES)
University of Calabria, Italy}
\IEEEauthorblockA{\IEEEauthorrefmark{2}francesco.feola@umu.se
\IEEEauthorrefmark{2}paolo.soda@umu.se, \IEEEauthorrefmark{4}p.soda@unicampus.it} 
\IEEEauthorrefmark{1}sarita.mourya@studenti.unicz.it
\IEEEauthorrefmark{3}pierangelo.veltri@dimes.unical.it  \\
\textsuperscript{1}These authors share first authorship.}


\maketitle
\begin{abstract}
Computed tomography (CT) and positron emission tomography (PET) provide complementary anatomical and functional information for cancer diagnosis and treatment planning. 
However, the widespread use of PET is limited by high radiation exposure, elevated costs, and restricted availability. 
To address these limitations, deep learning-based CT-to-PET translation has emerged as a promising approach for synthesizing PET-like information directly from CT images, although accurately modeling the large cross-modal gap remains challenging.
In this work, we propose a 3D CT-to-PET translation framework based on latent Brownian Bridge Diffusion (BBDM).
The method consists of two stages.
First, a Variational Autoencoder (VAE) is trained on paired CT–PET patches, integrating contrastive learning to improve latent alignment between anatomical and metabolic representations. 
Second, a BBDM is trained in the latent space to translate CT latent representations into their corresponding PET counterparts. 
The translated PET latents are then decoded and stitched to reconstruct the final 3D PET volume.
We evaluate the proposed approach on two publicly available datasets.
Quantitative results based on image fidelity and lesion-level PET-specific metrics demonstrate improved performance compared with competing methods. 
In particular, the proposed approach improves PET signal fidelity, better preserves clinically relevant uptake patterns, and shows improved performance in preserving small-lesion metabolic activation, paving the way for virtual imaging applications. Code is available at: \url{https://github.com/arco-group/3D-CT2PET-via-Latent-Brownian-Bridge-Diffusion}.
\end{abstract}
\begin{IEEEkeywords}
CT-to-PET Translation;
Brownian Bridge Diffusion;
Medical Image Synthesis;
Lesion-aware Image Translation
\end{IEEEkeywords}

\section{Introduction}
Computed Tomography (CT) and Fluorodeoxyglucose Positron Emission Tomography (FDG-PET, shortly PET in the following) provide complementary anatomical and functional information that is central to oncological diagnosis and treatment planning.
While CT captures structural abnormalities, PET measures metabolic activity and has shown improved diagnostic accuracy for the characterization of malignant lesions, particularly in lung cancer screening and staging~\cite{garcia2016assessment, shim2005non, veronesi2015positron}.
Despite its clinical value, PET remains limited by higher radiation exposure, elevated operational costs, and restricted availability, especially in low-resource settings where access to PET imaging is still scarce~\cite{who_pet_2022}. 
These limitations have motivated growing interest in computational approaches capable of estimating functional information directly from routinely acquired anatomical imaging while reducing both radiation burden and economic barriers. 
In this context, deep learning-based image-to-image translation has recently emerged as a promising framework for CT-to-PET synthesis, where a model learns a mapping from a source domain $X$ to a target domain $Y$~\cite{pang2021image}. 
Existing studies suggest that CT and PET share latent relationships between anatomical structures and metabolic activity~\cite{ esfahani2022pet}, supporting the feasibility of learning-based cross-modal translation.
Synthetic PET should therefore not be interpreted as a replacement for acquired PET, but rather as a probabilistic estimate of plausible metabolic activity inferred from CT data. 
Such synthesized information may provide complementary functional cues to support downstream clinical and computational workflows, including lesion analysis, patient triage, and computer-aided diagnosis.

Early CT-to-PET translation approaches mainly relied on Generative Adversarial Networks (GANs)~\cite{salehjahromi2024synthetic, Virtual-Scanner}, demonstrating the potential of deep generative models for PET synthesis.
However, GAN-based approaches often struggle to model the highly complex relationship between structural and functional information, leading to unstable training dynamics, hallucinated uptake patterns, and limited anatomical consistency~\cite{ben2019cross}. 
Moreover, several existing methods operate on 2D 
slices~\cite{Xu2025WholeBodyCTtoPET},\cite{Nguyen2025}, neglecting the inherently volumetric nature of PET/CT acquisitions and limiting their ability to capture inter-slice dependencies and whole-body contextual information.
More recently, Diffusion Models (DMs)~\cite{ddpm} have shown strong performance in image synthesis tasks by progressively denoising corrupted samples through an iterative stochastic process.
Compared with GANs, DMs generally provide improved generation stability and higher fidelity reconstructions, making them attractive for medical imaging applications. 
To mitigate the increased computational cost, latent diffusion models (LDMs)\cite{ldm} perform the denoising process in a low-dimensional latent-space, making them more practical for 3D medical imaging data.
Given a source image $x \in X$ and a target image $y \in Y$, most diffusion-based frameworks formulate image-to-image translation as conditional generation, modeling $p(y \mid x)$. While effective in some settings, this formulation lacks an explicit mapping between source and target domains and often struggles to bridge large distributional gaps between distinct imaging modalities.
To address this limitation, the Brownian Bridge Diffusion Model (BBDM)~\cite{inproceedings} formulates image-to-image translation as a stochastic Brownian Bridge process with boundary conditions defined by the source image $x$ and the target image $y$.
A recent study explored BBDM for CT-to-PET translation using 2D slice-based processing~\cite{Nguyen2025}, showing promising results for PET synthesis.

In this work, we propose X-Bridge, a 3D CT-to-PET translation approach based on latent Brownian Bridge diffusion. 
We first train a variational autoencoder (VAE) on paired CT/PET patches and integrate contrastive learning to improve latent alignment between anatomical and metabolic representations. 
Then, we train a BBDM in the latent space to translate CT latent representations into the corresponding PET latent targets. 
By operating directly in a 3D latent representation space, our approach captures volumetric contextual information while reducing the computational burden associated with full-resolution diffusion modeling.
Experimental results on two publicly available PET/CT datasets demonstrate improved quantitative performance and better preservation of small-lesion metabolic activation compared with competing methods.

\section{Methods}
\label{sec:methods}
Let $\mathbf{x} \in \mathbb{R}^{h \times w \times d}$ denote a CT volume and $\mathbf{y} \in \mathbb{R}^{h \times w \times d}$ the corresponding PET volume, where $h$, $w$, and $d$ denote the spatial dimensions.
The goal of CT-to-PET translation is to learn a mapping
\begin{equation}
\hat{\mathbf{y}} = f(\mathbf{x}) \approx \mathbf{y},
\end{equation}
where $f(\cdot)$ estimates PET metabolic activity from anatomical CT information.
To address the computational complexity associated with volumetric medical imaging, we adopt a patch-based formulation. Cubic 3D patches are extracted from paired CT/PET volumes using a sliding-window strategy and processed independently during training and inference.

\begin{figure*}
       \centering
        \includegraphics[width=1\linewidth, height=0.25\textheight]{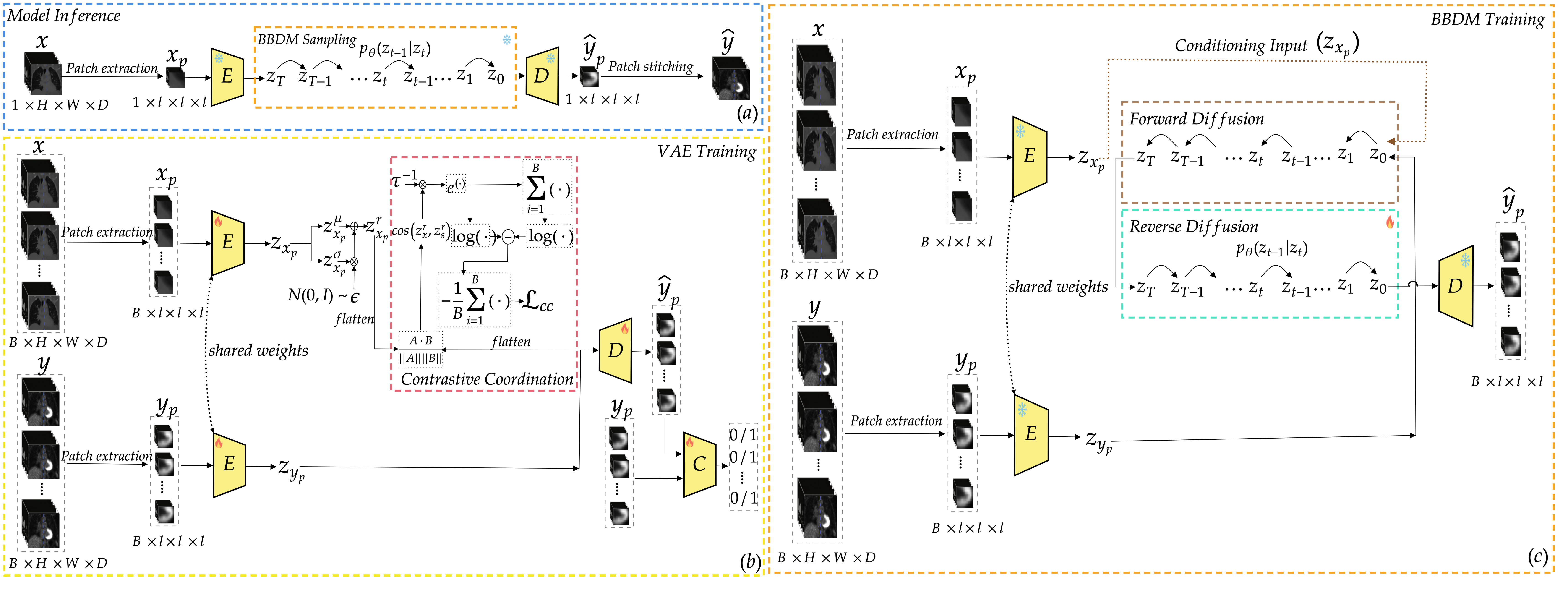}
\caption{Overview of the proposed X-Bridge approach. (a) Inference pipeline. CT patches are encoded into latent representations, translated into PET latent embeddings using BBDM, decoded, and stitched to reconstruct the target PET volume. (b) VAE training. Paired CT and PET patches are encoded using a shared encoder, while contrastive latent alignment and reconstruction objectives are jointly optimized. (c) BBDM training. The pretrained VAE encoder is frozen and used to project CT and PET patches into the latent space, where BBDM learns the conditional CT-to-PET translation process.}
    \label{fig:pipeline} 
\end{figure*}

As shown in Fig.~\ref{fig:pipeline} (a), our approach consists of two main components: a variational autoencoder (VAE) and a Brownian Bridge Diffusion Model (BBDM).
First, we train the VAE to learn a latent representation of CT and PET patches while contrastive learning promotes cross-modal latent alignment (Fig.~\ref{fig:pipeline} (b)). 
Second, we train the BBDM in the latent space of the VAE to translate CT latent embeddings into the corresponding PET latent representations (Fig.~\ref{fig:pipeline} (c)).
During inference, translated latent patches are decoded and stitched together to reconstruct the final synthetic PET volume.

\label{sec:inference_pipeline}

Given an input CT volume $\mathbf{x}$, we extract cubic patches
$\mathbf{x}_p \in \mathbb{R}^{l \times l \times l}$
using a sliding-window strategy with patch size $l$ and stride $s$:
\begin{equation}
\mathbf{x}_p = \mathbf{x}(i:i+l,\; j:j+l,\; k:k+l),
\end{equation}
where $i$, $j$, and $k$ denote the starting indices along each spatial axis and are incremented with stride $s$, i.e.,
$i,j,k \in \{0,s,2s,\dots\}$.
When $s < l$, adjacent patches overlap by $l-s$ voxels, enabling smoother transitions during reconstruction. 
After translation, all synthetic PET patches are stitched together to reconstruct the final PET volume $\hat{\mathbf{y}}$. 
Overlapping regions are averaged to reduce boundary artifacts and improve spatial consistency.

\subsection{Latent Representation Learning}
\label{VAE_training}

Fig.~\ref{fig:pipeline}-(b) illustrates the proposed VAE framework~\cite{VAE}, composed of an encoder $\mathcal{E}$ and a decoder $\mathcal{D}$. Given paired CT and PET patches $\mathbf{x}_p$ and $\mathbf{y}_p$
are processed using the same encoder branch with shared weight,
which maps both modalities into a latent space:
\begin{equation}
z_{x_p} = \mathcal{E}(\mathbf{x}_p),
\qquad
z_{y_p} = \mathcal{E}(\mathbf{y}_p).
\end{equation}

To regularize the latent space, we employ the Kullback--Leibler (KL) divergence loss:
\begin{equation}
\mathcal{L}_{\text{KL}}
=
D_{\text{KL}}
\left(
q(z|\mathbf{x})
\;\|\;
\mathcal{N}(0,\mathbf{I})
\right).
\end{equation}

Since CT and PET encode substantially different anatomical and metabolic information, their latent representations are not naturally aligned. To reduce this modality gap, we employ an InfoNCE-based contrastive objective between paired CT and PET latent embeddings:
\begin{equation}
\mathcal{L}_{\text{cc}} = -\frac{1}{B} \sum_{m=1}^{B} \log \left( 
\frac{ \ e^{\left({ \text{cos}(z_{x_p}^{(m)}, z_{y_p}^{(m)}) }/{\tau} \right)} }
{ \sum_{n=1}^{B} \ e^{\left({ \text{cos}(z_{x_p}^{(m)}, z_{y_p}^{(n)}) }/{\tau} \right)} }
\right)
\end{equation}
where $B$ denotes the batch size, $\text{cos}(\cdot,\cdot)$ is the cosine similarity, and $\tau$ is the temperature parameter. This objective encourages aligned CT/PET latent embeddings while separating unrelated samples, facilitating the subsequent latent translation process.
The decoder $\mathcal{D}$ reconstructs PET patches from the latent representation:
\begin{equation}
\hat{\mathbf{y}}_p = \mathcal{D}(z_{y_p}).
\end{equation}

To preserve reconstruction fidelity, we employ a $\mathcal{L}_1$ reconstruction loss:
\begin{equation}
\mathcal{L}_{\text{recon}}
=
\|
\mathbf{y}_p - \hat{\mathbf{y}}_p
\|_1.
\end{equation}
Additionally, we employ a perceptual loss~\cite{Johnson2016} to preserve high-level structural information and a patch-based adversarial loss~\cite{Isola2017} to improve perceptual realism and fine-grained details.
The overall training objective is defined as:
\begin{equation}
\mathcal{L}_{\text{T}}
=
\boldsymbol{\alpha}^{\top}\mathbf{L},
\end{equation}
where
\begin{equation}
\boldsymbol{\alpha}
=
[1,\beta,\gamma,\delta,\lambda]^\top
\end{equation}
and
\begin{equation}
\mathbf{L}
=
\begin{bmatrix}
\mathcal{L}_{\text{recon}} \\
\mathcal{L}_{\text{KL}} \\
\mathcal{L}_{\text{adv}} \\
\mathcal{L}_{\text{percep}} \\
\mathcal{L}_{\text{cc}}
\end{bmatrix}.
\end{equation}
Here, $\beta$, $\gamma$, $\delta$, and $\lambda$ control the contribution of each loss term.
\subsection{Latent Brownian Bridge Diffusion}
\label{sec:BBDM}
Fig.~\ref{fig:pipeline}-(c) illustrates the proposed Brownian Bridge Diffusion Model (BBDM)~\cite{inproceedings}. Building on the latent representation learning described in Section~\ref{VAE_training}, the diffusion process operates directly in the latent space using the CT latent representation $z_{x_p}$ and the corresponding PET latent representation $z_{y_p}$ extracted by the pretrained VAE encoder, whose parameters are frozen during diffusion training.
Unlike conventional conditional diffusion models, BBDM explicitly models a stochastic bridge between source and target distributions. In our setting, the diffusion trajectory is constrained by the CT latent embedding $z_{x_p}$ and the target PET latent embedding $z_{y_p}$, enabling a progressive translation between anatomical and metabolic representations.
To improve computational efficiency, the diffusion process is performed in the compressed latent space rather than on full-resolution images. Following~\cite{inproceedings}, the forward Brownian bridge process is defined as:
\begin{equation}
q_{\text{BB}}(z_t \mid z_{y_p}, z_{x_p})
=
\mathcal{N}
\left(
z_t;
(1-m_t)z_{y_p}
+
m_t z_{x_p},
\delta_t \mathbf{I}
\right),
\end{equation}
where $m_t$ controls the interpolation between source and target latent representations and $\delta_t$ defines the variance schedule.

The diffusion model is trained using the Brownian Bridge noise prediction objective:
\begin{equation}
\mathbb{E}_{z_{y_p},z_{x_p},\epsilon}
\left[c_{\epsilon t}
\left\|
m_t(z_{x_p}-z_{y_p})
+
\sqrt{\delta_t}\epsilon
-
\epsilon_{\theta}(z_t,t)
\right\|^2
\right]
\end{equation}
where  $c_{\epsilon t}$ denotes a timestep-dependent weighting coefficient, typically defined based on the Brownian Bridge variance term $\delta_t$. $\epsilon \sim \mathcal{N}(0,\mathbf{I})$ and $\epsilon_{\theta}$ denotes the noise prediction network.
During inference, the trained BBDM progressively translates CT latent embeddings into PET latent representations, which are subsequently decoded by the VAE decoder to reconstruct the synthetic PET patches. To accelerate inference while preserving generation quality, we adopt the DDIM sampling strategy~\cite{DDIM}.
\section{Materials}
\textbf{Internal dataset.} Experiments were conducted using the FDG-PET-CT-Lesions dataset~\cite{Gatidis2022FDGPETCTLesions}, a publicly available collection of 1,014 whole-body FDG-PET/CT scans acquired at the University Hospital Tübingen between 2014 and 2018. 
The dataset includes subjects diagnosed with malignant lymphoma, non-small cell lung cancer, and malignant melanoma, as well as PET-negative control subjects without evidence of malignant findings.
The dataset was split into a training set comprising 864 paired CT–PET scans and an internal test set of 150 paired scans, guaranteeing that patches extracted from a patient study are only in the training or test set.
\textbf{External dataset.} External validation was performed using the ENHANCE.PET 1.6k dataset~\cite{ENHANCE_PET_MOOSE_1_6k}, a publicly available multi-center collection of whole-body FDG-PET/CT scans acquired across different institutions and scanner vendors. 
The dataset comprises 1,579 paired PET/CT studies collected from subjects with lung cancer, lymphoma, melanoma, and PET-negative controls.
Since data from the University Hospital Tübingen overlap with the internal dataset used in this study, these cases were excluded from external validation to avoid data leakage. The final external cohort consists of 579 paired PET/CT scans acquired at the University Hospital Leipzig and Azienda Ospedaliero Universitaria Careggi.

\textbf{Preprocessing.} CT volumes were resampled to match the spatial resolution and voxel spacing of the corresponding PET scans using trilinear interpolation. Rigid registration was subsequently applied to correct residual misalignments caused by patient motion or acquisition differences.
PET intensities were converted to standardized uptake values (SUVs) and clipped to the range $[0,20]$. Finally, both CT and PET volumes were normalized to the range $[0,1]$ using min-max normalization.
In this work, we focus on the lung region. Lung masks were automatically extracted using the MOOSE segmentation framework~\cite{shiyamsundar2022fully} and used to isolate the lung volumes for model training and evaluation.
\section{Experimental setup}
To evaluate X-Bridge, we compare our approach against three competing methods: a 3D GAN-based CT-to-PET translation approach~\cite{Virtual-Scanner}, a 3D latent diffusion model (LDM), and CPDM~\cite{Nguyen2025}, which combines 2D VQGAN latent representations with Brownian Bridge diffusion for slice-wise CT-to-PET translation.
\subsection{Implementation details}
Following the patch-based strategy described in Section~\ref{sec:inference_pipeline}, cubic patches of size $32 \times 32 \times 32$ were extracted using a sliding window with a stride of 16.
The VAE was trained using the Adam optimizer with a learning rate of $1 \times 10^{-4}$ and a batch size of 128 for 300 epochs. 
The training objective $\mathcal{L}_{\text{T}}$ combines the different loss terms using $\beta = 1 \times 10^{-7}$, $\gamma = 0.1$, $\delta = 0.3$, and $\lambda = 0.05$.
The diffusion model was subsequently trained using the Adam optimizer with a learning rate of $1 \times 10^{-4}$ and a batch size of 32 for 300 epochs. 
The number of diffusion timesteps was set to 1000, while 200 DDIM denoising steps were used during inference.
All experiments were performed on a NVIDIA A40 GPU. 
The source code is available \href{https://github.com/arco-group/3D-CT2PET-via-Latent-Brownian-Bridge-Diffusion}{here}.

\subsection{Evaluation Metrics}
We used two complementary quantitative analyses: standard image quality metrics and PET-specific signal fidelity metrics based on standardized uptake values (SUVs).
\subsubsection*{Standard Image Quality Assessment Metrics}
We used the Peak Signal-to-Noise Ratio (PSNR) and the Structural Similarity Index Measure (SSIM) \cite{DiFeola2023}.
PSNR assesses image quality based on the ratio between the maximum possible pixel intensity ($MAX_{\hat{y}}$) and the mean squared error (MSE) between $\hat{y}$ and $y$ 
while the SSIM evaluates the perceptual similarity between two images by considering their luminance, contrast, and structural information, ranging from 0 to 1.
\subsubsection*{PET signal fidelity}
To assess the clinical reliability of synthetic PET images, we evaluated the agreement between PET-derived quantitative biomarkers computed on synthetic and ground-truth PET volumes. Specifically, we considered the maximum standardized uptake value ($SUV_{max}$), mean standardized uptake value ($SUV_{mean}$), metabolic tumor volume ($MTV_{\alpha}$), and total lesion glycolysis ($TLG_{\alpha}$).
Let $\mathcal{R} \subset \mathbf{y}$ denote a region of interest (ROI) within the PET volume $\mathbf{y}$. The uptake metrics are defined as:
\begin{equation}
SUV_{\text{max}}
=
\max_{v \in \mathcal{R}} y(v),
\qquad
SUV_{\text{mean}}
=
\frac{1}{|\mathcal{R}|}
\sum_{v \in \mathcal{R}} y(v),
\end{equation}
where $|\mathcal{R}|$ denotes the number of voxels in the ROI.

The metabolic tumor volume is computed as:
\begin{equation}
MTV_{\alpha}
=
|\mathcal{R}_{\alpha}|,
\qquad
\mathcal{R}_{\alpha}
=
\{v \in \mathcal{R} \mid y(v) > \alpha\},
\end{equation}
while total lesion glycolysis is defined as:
\begin{equation}
TLG_{\alpha}
=
\left(
\frac{1}{|\mathcal{R}_{\alpha}|}
\sum_{v \in \mathcal{R}_{\alpha}} y(v)
\right)
\cdot
MTV_{\alpha}.
\end{equation}
To further evaluate lesion-level uptake preservation, we computed the Dice score between thresholded synthetic and ground-truth PET lesion masks:
\begin{equation}
Dice_{\alpha}
=
\frac{
2|M_{\alpha}^{syn} \cap M_{\alpha}^{gt}|
}{
|M_{\alpha}^{syn}| + |M_{\alpha}^{gt}|
},
\end{equation}
where $M_{\alpha}^{syn}$ and $M_{\alpha}^{gt}$ denote the binary lesion masks obtained by thresholding the synthetic and ground-truth PET volumes at SUV threshold $\alpha$.

We considered two absolute uptake thresholds, $\alpha=1.5$ and $\alpha=2.5$, commonly used to characterize low-uptake and malignant lesions~\cite{fletcher2010pet,li2019prognostic}.
\section{Results and Discussion}
\label{results}
This section presents an in-depth evaluation of X-Bridge through both quantitative and qualitative analyses.
In CT-to-PET translation, lesion uptake regions typically occupy only a small fraction of the whole image volume, making the preservation of small lesions and high-intensity metabolic patterns particularly challenging. Therefore, we first focused on a lesion-level analysis to evaluate the ability of each model to preserve lesion morphology and metabolic activity.
\begin{figure}[h]
    \centering
    \includegraphics[width=1.0\linewidth, height=0.2\textheight]{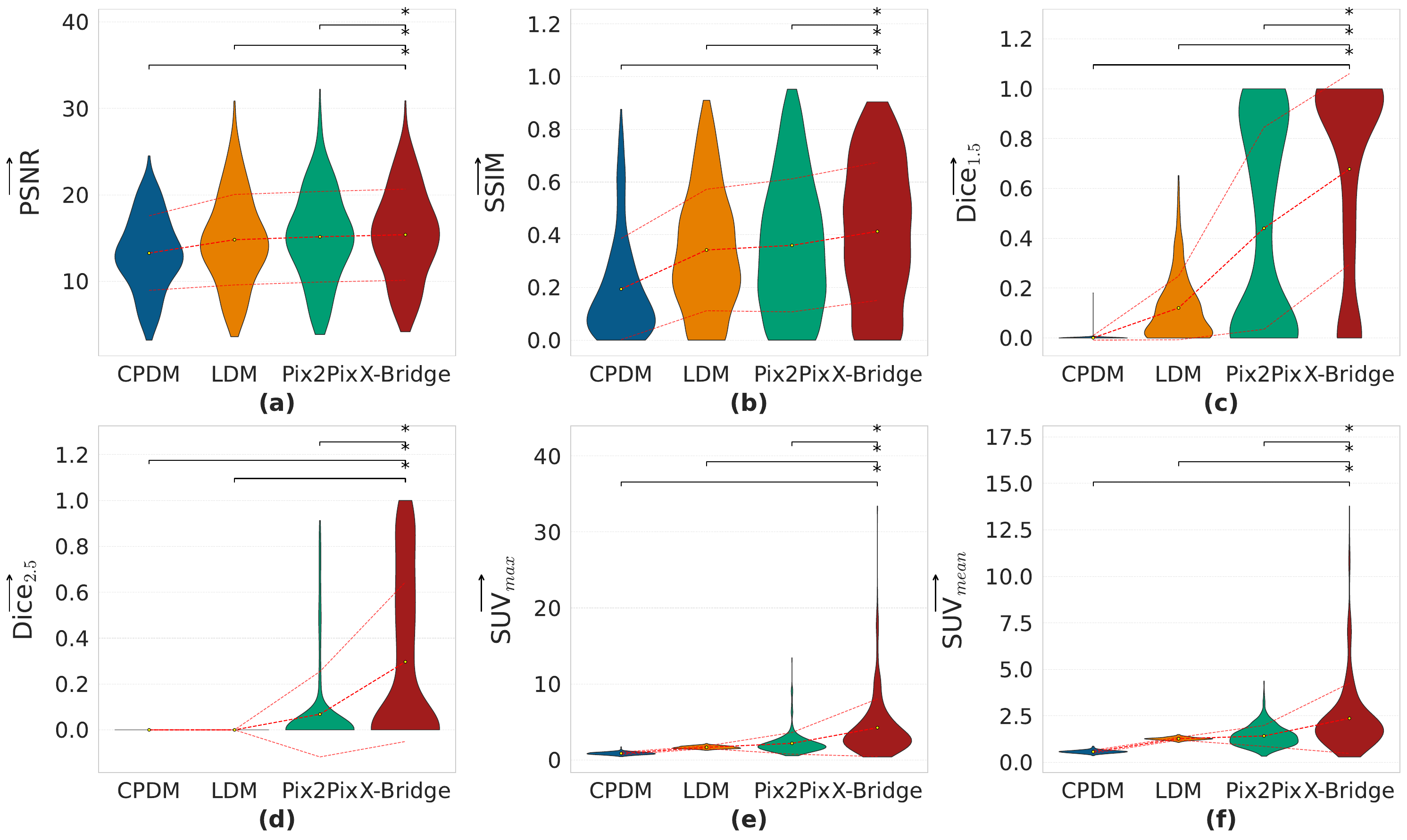}
    \caption{Lesion-level comparison of PET volumes synthesized by CPDM, LDM, Pix2Pix, and X-Bridge. Violin plots report (a) PSNR, (b) SSIM, (c) Dice$_{1.5}$, (d) Dice$_{2.5}$, (e) SUV$_{max}$, and (f) SUV$_{mean}$. Violin width indicates sample density, yellow markers denote medians, and red dashed/dotted lines represent mean $\pm$ standard deviation. Asterisks indicate statistically significant differences ($p<0.05$) between X-Bridge and competing methods using the Wilcoxon signed-rank test with Bonferroni correction.}
    \label{fig:Lesion_level_with_stat}
\end{figure}
\subsection{Lesion-level analysis}
\label{sec:lesion}
\begin{figure}[t]
    \centering
    \includegraphics[width=0.7\linewidth]{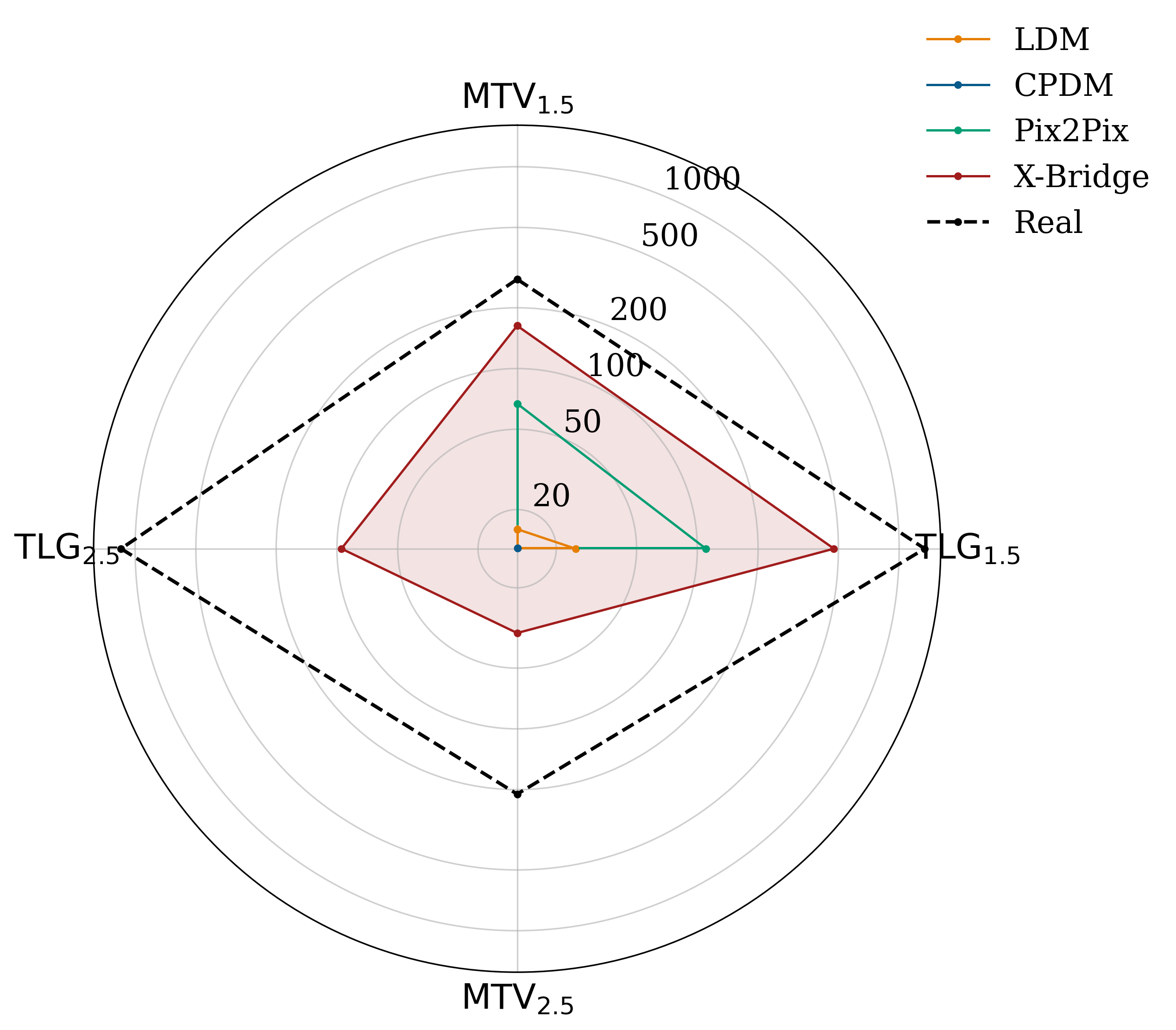}
    \caption{Radar comparison of median lesion-level MTV and TLG measured at SUV thresholds 1.5 and 2.5 for LDM, CPDM, Pix2Pix, and X-Bridge, shown on a logarithmic radial scale. 
     Closer agreement with the real-value distribution indicates better preservation of metabolically active lesion extent and uptake burden}
    \label{fig:mtv_tlg_plot}
\end{figure}

\begin{figure}[t]
    \centering
    \includegraphics[width=\linewidth]{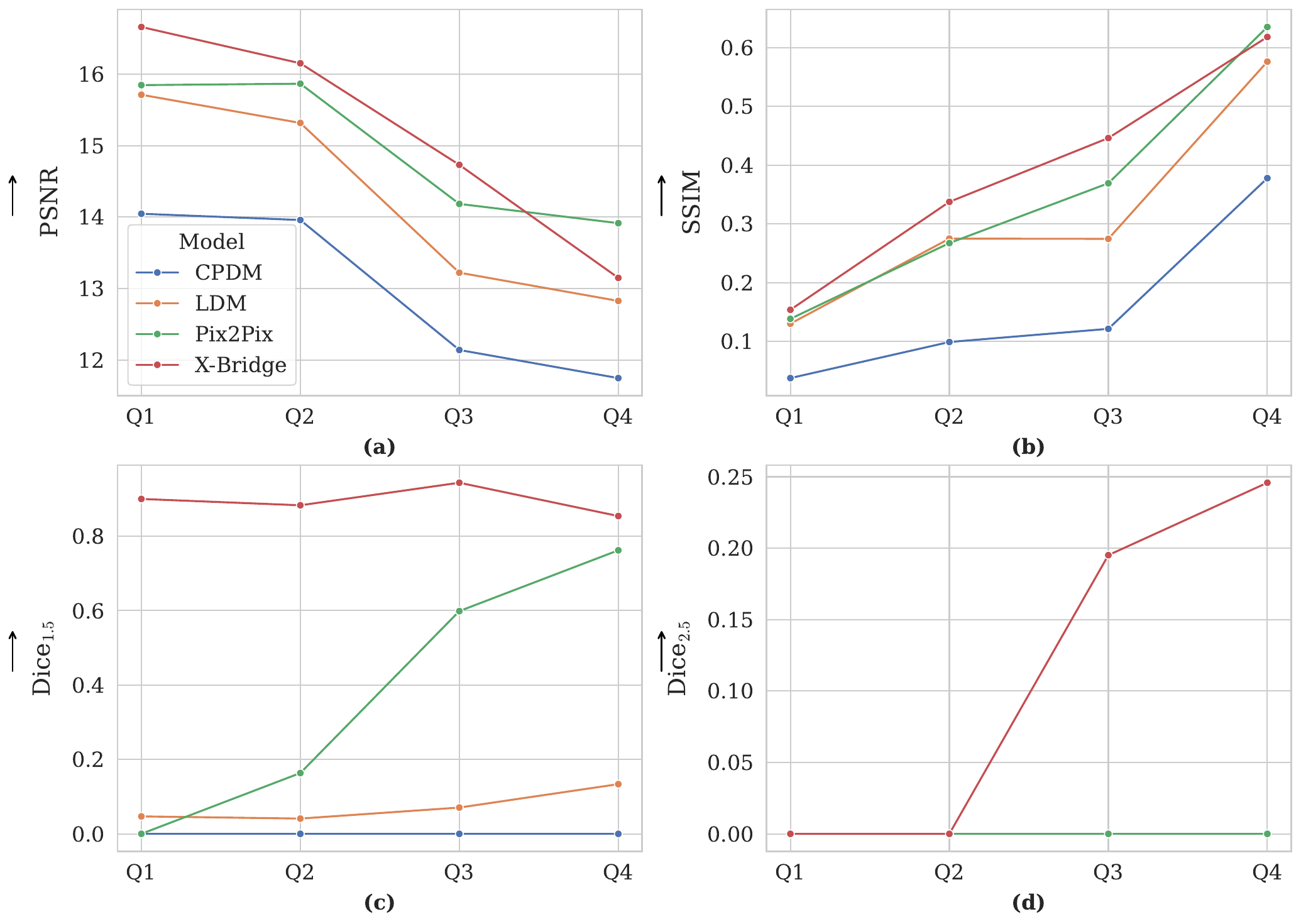}
    \caption{Lesion-size analysis of generated PET volumes. Plots report lesion-level (a) PSNR, (b) SSIM, (c) Dice$_{1.5}$, and (d) Dice$_{2.5}$ stratified by lesion-size quartiles: small (Q1), medium (Q2), large (Q3), and very large (Q4), for CPDM, LDM, Pix2Pix, and X-Bridge.}
    \label{fig:Lesion_by_size_group}
\end{figure}

Fig.~\ref{fig:Lesion_level_with_stat}(a--f) reports lesion-level results using image reconstruction, lesion overlap, and PET uptake metrics.
X-Bridge achieves higher PSNR, SSIM, Dice$_{1.5}$, Dice$_{2.5}$, SUV$_{max}$, and SUV$_{mean}$ values compared to competing methods, showing improved lesion generation fidelity and better preservation of metabolic uptake patterns.
In particular, the higher Dice$_{1.5}$ and Dice$_{2.5}$ values indicate a more accurate reconstruction of metabolically active lesion areas and improved preservation of high-uptake lesion regions.
Fig.~\ref{fig:mtv_tlg_plot} further supports these findings, showing that X-Bridge achieves substantially higher MTV and TLG values at both SUV thresholds (1.5 and 2.5). 
In contrast, the competing methods exhibit lower values, indicating a reduced preservation of metabolically active lesion volume and uptake burden during PET synthesis.
Turning to Fig.~\ref{fig:Lesion_by_size_group}, we report lesion-level performance stratified by lesion-size quartiles using PSNR, SSIM, Dice$_{1.5}$, and Dice$_{2.5}$. 
X-Bridge  achieves improved lesion synthesis fidelity across all lesion-size groups, with the largest advantage observed for small lesions.
PSNR decreases for all approaches as lesion size increases, reflecting the greater complexity of translating larger lesion regions. 
Nevertheless, X-Bridge maintains the highest PSNR across most lesion-size groups. 
SSIM generally improves for larger lesions due to the increased spatial context available during translation, while X-Bridge  achieves higher structural similarity, including for small lesions where preserving lesion morphology is particularly challenging.

\begin{figure}[h]
    \centering
    \includegraphics[width=\linewidth]{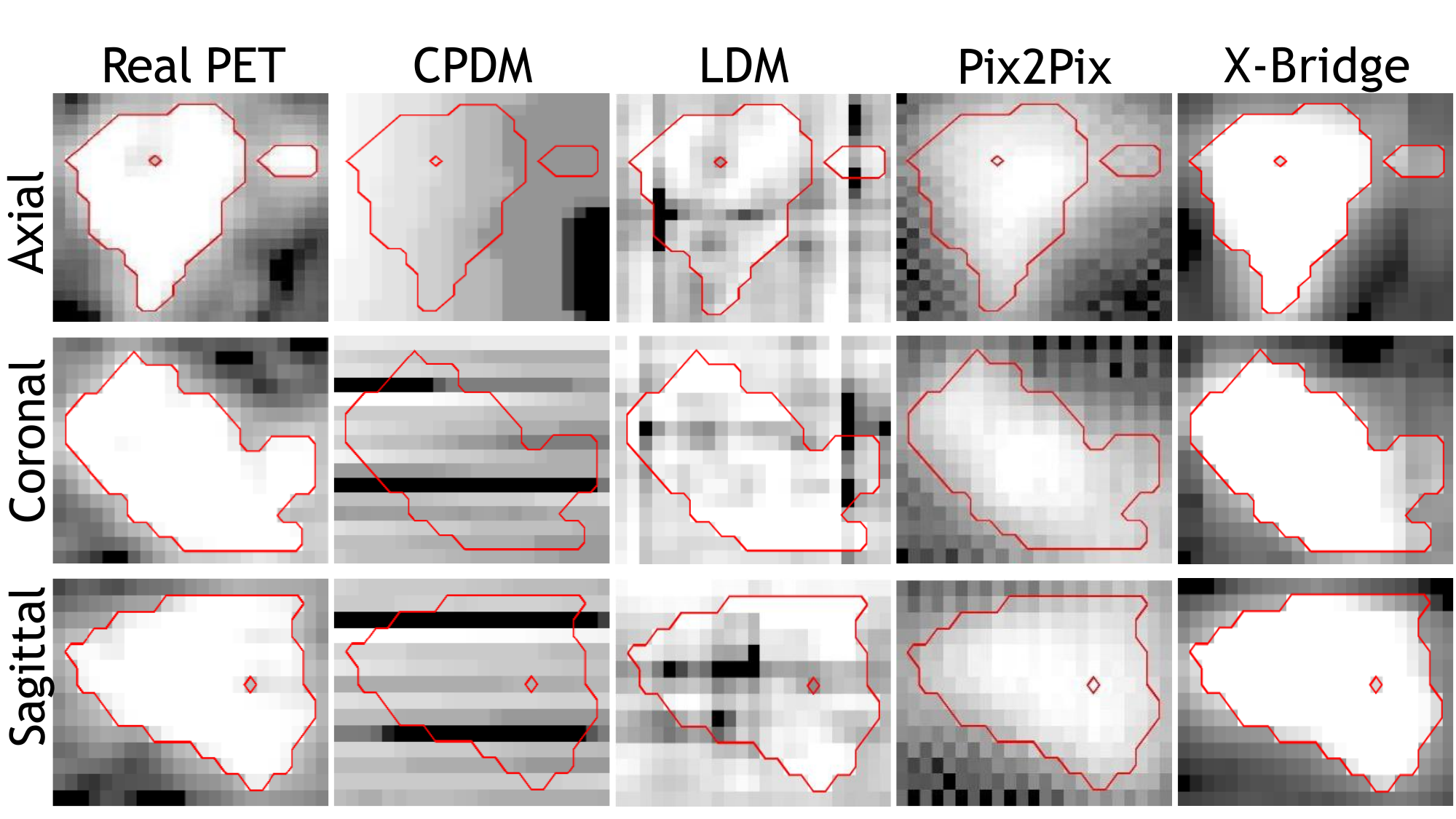}
    \caption{Qualitative comparison of Real PET, LDM, CPDM, Pix2Pix and X-Bridge, shown from left to right in axial, coronal, and sagittal views with lesion boundaries overlaid in red.}
    \label{fig:lesion_zoomed}
\end{figure}

Dice-based metrics further highlights X-Bridge's advantage. Dice$_{1.5}$ values consistently remains above 0.8 across all quartiles, indicating stable preservation of metabolically active regions, while competing methods achieves substantially lower overlap, especially in the lower quartiles. At the stricter Dice$_{2.5}$ threshold,all methods decline because compact high-uptake regions are harder to preserve; however, X-Bridge remains the only approach achieving meaningful overlap for medium-to-large lesions, whereas the competing methods remain close to zero across most quartiles.
In Fig.~\ref{fig:lesion_zoomed}, we report qualitative examples, showing axial, coronal, and sagittal views of lesion regions together with the annotated lesion contours. Compared with competing methods, X-Bridge better preserves lesion shape, uptake distribution, and boundary consistency.
In particular, the 2D slice-based CPDM exhibits substantial inconsistencies across views, highlighting its limited ability to preserve coherent 3D lesion-level information.
\begin{table}[!t]
\centering
\caption{Whole-lung image quality evaluation and external validation across models. Internal and external rows report mean PSNR and SSIM, and the last two rows show the relative difference: $\Delta(\%) = 100 \times (\mathrm{Avg}_{\mathrm{Internal}}-\mathrm{Avg}_{\mathrm{External}})/\mathrm{Avg}_{\mathrm{External}}$.}
\label{tab:external_validation}
\scriptsize
\setlength{\tabcolsep}{3pt}
\renewcommand{\arraystretch}{1.05}
\begin{tabular}{llcccc}
\hline
\textbf{Dataset} & \textbf{Metric} & \textbf{CPDM} & \textbf{LDM} & \textbf{Pix2Pix} & \textbf{X-Bridge} \\
\hline
Internal & PSNR & 26.7 & 26.3 & 33.30 & 29.99 \\
         & SSIM & 0.810 & 0.580 & 0.849 & 0.779 \\
External & PSNR & 26.6 & 26.1 & 29.40 & 29.91 \\
         & SSIM & 0.804 & 0.600 & 0.718 & 0.802 \\
$\Delta$ & PSNR (\%) & 0.38 & 0.77 & 13.27 & 0.27 \\
          & SSIM (\%) & 0.87 & -3.33 & 18.25 & 2.87 \\
\hline
\end{tabular}
\end{table}
\subsection{External validation and volume-level performance}
Table~\ref{tab:external_validation} reports volume-level PSNR and SSIM on the internal and external datasets; lesion-level evaluation was not possible externally because annotations were unavailable. X-Bridge outperforms LDM and CPDM across both datasets and performs comparably to Pix2Pix. It achieves the highest external PSNR (29.91), while Pix2Pix records slightly higher internal PSNR and SSIM.
Relative variation analysis further demonstrates X-Bridge’s cross-dataset robustness, with the smallest PSNR variation ($\Delta=0.27\%$), compared with larger shifts for CPDM and Pix2Pix. Although CPDM shows stable SSIM across datasets, its absolute performance remains below that of X-Bridge and Pix2Pix, whereas Pix2Pix exhibits the largest SSIM degradation ($18.25\%$), indicating lower robustness to distribution shifts. Overall, X-Bridge offers a favorable balance between global image fidelity and cross-dataset generalization.
\section{Conclusion}
In this work, we proposed X-Bridge, a 3D CT-to-PET translation approach that combines contrastive latent representation learning with Brownian Bridge diffusion.
The experimental results showed that the proposed approach preserves smaller lesions and compact high-uptake regions more effectively compared to existing methods based on GAN and diffusion.
This is particularly relevant in CT-to-PET translation, where metabolically active lesion regions occupy only a small and highly imbalanced fraction of the whole image volume.
Lesion-level analyses demonstrated that X-Bridge achieves improved preservation of clinically relevant metabolic activity across SUV-, Dice-, MTV-, and TLG-based metrics, while maintaining competitive global image fidelity. 
External validation further confirmed the robustness of the proposed latent Brownian Bridge formulation under cross-dataset evaluation.
Despite promising results, some limitations remain. First, our sliding-window reconstruction uses uniform averaging in overlapping patches, which may cause minor discontinuities when adjacent predictions differ. Future work will explore adaptive blending to improve smoothness while preserving local details. secondly, evaluation is currently limited to the lungs regions: broader validation is needed in other organs and whole-body imaging, where structural and metabolic properties may differ.

We will also extend X-Bridge with long-tail generative learning to improve synthesis of rare lesion patterns and small high-uptake regions. We will also explore uncertainty-aware CT-to-PET translation methods to account the task's inherently under-constrained nature, as well as bridge-based diffusion methods for multimodal conditional generation across complementary imaging modalities \cite{Molino2026}.

\section*{Acknowledgment}
Resources are provided by the National Academic Infrastructure for Supercomputing in Sweden (NAISS) and the Swedish National Infrastructure for Computing (SNIC) at Alvis @ C3SE, partially funded by the Swedish Research Council through grant agreements no. 2022-06725 and no. 2018-05973. 
This work was partially supported by: i) Project PNRR-MCNT2-2023-12377755 (CUP: C83C24000220007) - Rafforzamento e potenziamento della ricerca biomedica del SSN, finanziato dall’Unione europea – NextGenerationEU, ii) Kempe Foundation project JCSMK24-0094, iii) University Campus Bio-Medico di Roma, project IDEA.

\end{document}